\documentclass[a4paper]{article}

\usepackage{INTERSPEECH2020}

\title{A Dual-Decoder Conformer for Multilingual Speech Recognition}
\name{Krishna D N}
\address{
  Freshworks Inc.
  }
\email{krishna.nanjappa@freshworks.com}

\begin{document}

\maketitle
\begin{abstract}
Transformer-based models have recently become very popular for sequence-to-sequence applications such as machine translation and speech recognition. This work proposes a dual-decoder transformer model for low-resource multilingual speech recognition for Indian languages. Our proposed model consists of a Conformer [1] encoder, two parallel transformer decoders, and a language classifier. We use a phoneme decoder (PHN-DEC) for the phoneme recognition task and a grapheme decoder (GRP-DEC) to predict grapheme sequence along with language information. We consider phoneme recognition and language identification as auxiliary tasks in the multi-task learning framework. We jointly optimize the network for phoneme recognition, grapheme recognition, and language identification tasks with Joint CTC-Attention [2] training. Our experiments show that we can obtain a significant reduction in WER over the baseline approaches. We also show that our dual-decoder approach obtains significant improvement over the single decoder approach.

\end{abstract} 

\noindent\textbf{Index Terms}: multilingual speech recognition, transformers, multi-task learning, ASR, low-resource

\section{Introduction}
Studies have shown that more than 22 main spoken languages in India, and apart from Hindi and Indian English, many other languages are considered low resources due to the scarcity of the data and the speaker population. However, many Indian languages have acoustic similarities, and using multilingual acoustic models can be a good fit for building better speech recognition models. In this work, we explore multilingual speech recognition models for Indian languages.

Recent advances in deep learning have shown significant improvements in the speech recognition field. Conventional speech recognition models use a combination of Gaussian Mixture Models(GMMs) with Hidden Markov Models (HMMs) or Deep neural networks (DNNs) with HMMs. Many hybrid approaches have been proposed in the past [5,6,7,8,9,10] for multilingual speech recognition, and these approaches use initial layers of the acoustic models as feature extractors for language-dependent layers during adaptation. Hybrid multilingual ASR models with multi-task learning have been proposed for multilingual speech recognition tasks [3, 12]. These models use auxiliary tasks to improve the overall system performance. Conventional or Hybrid models require components like an acoustic model, pronunciation dictionary, and a language model. These components may become hard to build for low resource languages.

End-to-End (E2E) approaches for speech recognition have become very popular in recent years [2,13]. Unlike hybrid models, E2E models do not require a pronunciation dictionary or alignments to train. In the case of E2E speech recognition, acoustic models, pronunciation dictionaries, and language models are integrated into a single framework. Sequence-to-sequence [14] models are a special class of E2E models, which use encoder and decoder framework to learn a mapping from acoustic data to text. [15] proposed an attention-based E2E model called LAS using LSTM based encoder and decoder. [16] proposed to use the LAS model for multilingual speech recognition for 12 Indian languages, and they showed that training a single model for all the languages improves each language's performance. Using language identification as an auxiliary task for the LAS framework has been proposed in [17], and they show that it can improve the recognition performance for each language.
Recently, Transformers [18] have shown the state of the art performance for many Natural language processing problems [19]. The transformer models are being used in many areas of speech including speech recognition [1,20,21,22,31], emotion recognition [23], language translation [24] etc. Transformer-based approaches are explored in [4,16,17,25,26]. [1] proposed a variant of a transformer called the conformer model for speech recognition. The conformer model contains convolution layers augmentation inside the transformer layers. The convolution layers capture the local interaction of the features, where the self-attention layers capture the global feature interactions.

This paper proposes a dual-decoder conformer model for multilingual speech recognition with multi-task learning. Our model consists of a single conformer encoder, two transformer decoders called phoneme decoder (PHN-DEC) and grapheme decoder (GRP-DEC), and a language classifier. The phoneme decoder is trained to predict a sequence of phonemes, and the grapheme decoder is used to predict a sequence of grapheme units for a given utterance. The grapheme decoder first predicts the language label, followed by a sequence of grapheme units from a multilingual vocabulary. The vocabulary contains grapheme units from all the languages, and it is also augmented with language labels. The phoneme sequence prediction and language identification act as auxiliary tasks for the model during training. We use two different approaches to provide language information both at the decoder and the encoder. 1) We use a separate language classifier at the encoder to predict the language information for the given utterance. 2) The grapheme decoder is tasked to predict the language information before generating the letter sequence.
The entire model is trained in an end-to-end fashion using joint CTC-Attention [2] objective function. We conduct our experiments on six low-resource Indian language speech recognition data released by Microsoft and Navan Tech as part of a special session on "Multilingual and code-switching ASR challengees for low resource Indian languages in INTERSPEECH 2021". We show that our approach obtains significant improvements over the baseline approaches.

The organization of the paper is as follows. Section 2 explains our proposed approach in detail. In section 3, we give details of the dataset and experimentation. Finally, section 4 describes our results.

\section{Proposed approach}
Transformer-based E2E models have become more common for many speech applications, including speech recognition, emotion recognition, language identification, speaker identification, etc.
In this work, we propose a dual-decoder transformer model for low-resource multilingual speech recognition for 6 Indian languages. Our proposed model architecture is shown in Figure 1. It consists of a conformer Encoder, two transformer decoders, and a language classifier. The conformer encoder is similar to the transformer encoder, but the self-attention layers are augmented with convolution layers. The model contains two decoders, PHN-DEC and GRP-DEC, as shown in Figure 1. The PHN-DEC network consists of M cross-attention layers, and it is responsible for the phoneme recognition task. Whereas the GRP-DEC also consists of M cross-attention layers, it is responsible for predicting the grapheme sequence for the given input utterance. The language classifier has two linear layers followed by a softmax layer for language label prediction. The phoneme recognition task and language identification task are considered auxiliary tasks for the model training, and it helps to learn shared feature representation across languages and improves the overall system performance. On the other hand, the GRP-DEC is responsible for predicting the character sequence for a given utterance. The language information is provided at both encoder and decoder to make sure the model understands which characters should be chosen from the vocabulary while decoding. We pool the feature sequence from the last layer of the encoder network and feed it to the language classifier. In the decoder end, we ask the GRP-DEC decoder to predict the language label before generating the grapheme sequence. Since the model is trained with multilingual data, the vocabulary should consist of all the possible characters from all six Indian languages, such as Telugu, Tamil, Gujarati, Marathi, Hindi, and Odiya. The entire network is trained in an end-end-fashion using the joint CTC-Attention objective. The loss function consists of a weighted sum of CTC loss, phoneme recognition loss, and grapheme recognition loss.

\begin{figure}[t]
  \centering
  \includegraphics[width=\linewidth]{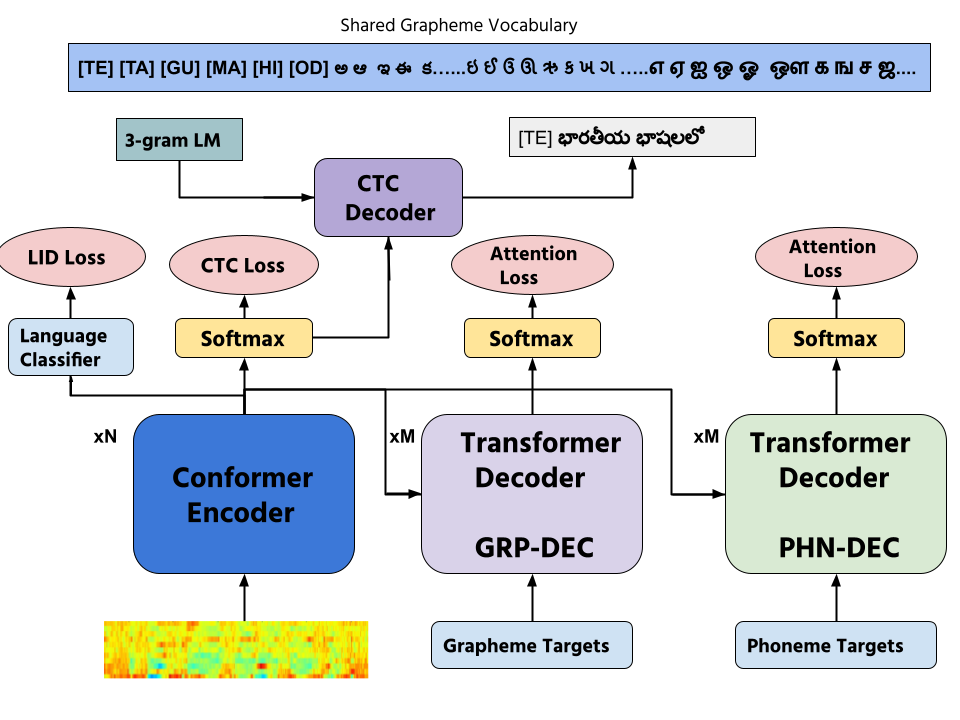}
  \caption{Proposed model architecture.}
  \label{fig:speech_production}
\end{figure}

\subsection{Conformer Encoder}
The conformer model is a variant of transformer architecture where the self-attention layers are augmented with convolution layers. The multi-head self-attention layers use a scaled dot-product attention mechanism to capture the global variations in the feature sequence. Since the convolution operation captures the local information, augmenting transformer layers with convolution layers improves the overall system performance. The conformer encoder consists of a convolution sub-sampling layer followed by N conformer blocks. Each conformer block consists of 2 feed-forward (Linear) layers, a convolution module, and a layer normalization layer. Processing longer feature sequences could be problematic for transformer architectures due to linear complexity in the self-attention layers. To avoid memory issues, we use an initial sub-sampling layer to reduce the input sequence length. The convolution sub-sampling reduces the frame rate by a factor of 4. The convolution module in the conformer block is responsible for learning local feature representation from feature sequences. The convolution module consists of a pointwise convolution layer with an expansion factor of 2 with a GLU activation layer followed by a 1-D Depthwise convolution layer, Batch normalization, and a swish activation. The conformer encoder takes a sequence of acoustic features as input and transforms it through multiple transformation layers using conformer blocks to learn high-level feature representation. The conformer encoder's output is used as input to the PHN-DEC, GRP-DEC, Language classifier, and CTC. The CTC layer predicts the language identity information in the beginning and predicts grapheme units for every frame. We use CTC decoding with a 3-gram multilingual language model during the inference stage. 

\subsection{Dual Decoder Network}
Recent end-to-end speech recognition models consist of an encoder-decoder framework where the encoder processes the audio features and the decoder generates a sequence of output units (characters or phonemes) conditioned on the encoder's output. In this paper, we propose to use two different decoders known as phoneme decoder (PHN-DEC) and grapheme decoder (GRP-DEC). The phoneme decoder consists of several transformer layers and takes conformer encoder output and phoneme embeddings as input, as shown in Figure 1. Each decoder layer in PHN-DEC consists of feed-forward layers and multi-head cross attention layers. The output of the PHN-DEC is a sequence of phoneme corresponding to the input utterance. We use 68 phonemes from the common phoneme label set provided by IndicTTS. The grapheme decoder (GRP-DEC) is also a transformer-based decoder consisting of M transformer layers. The GRP-DEC takes character embeddings as input along with conformer encoder last layer output to predict the sequence of grapheme units for a given input utterance. The vocabulary of the GRP-DEC is consists of all possible characters from 6 Indian languages. The vocabulary is also augmented with language labels, as shown in Figure 1.

\subsection{Language classifier}
Generally, end-to-end multilingual speech recognition models are provided language information either at the encoder or at the decoder. [4] explores various ways to inject language information into the transformer-based multilingual speech recognition model. In this work, we decide to provide the language information at both encoder and decoder. In the decoder case, we force the network to predict the language label before the sequence prediction, and in the case of the encoder, we use the encoder features as input to a language classifier. We first pool the features from the encoder's last layer using the Average adaptive pooling layer and feed the utterance-level feature to the language classifier, as shown in Figure 1. The language classifier consists of 2 linear layers followed by a softmax layer. It takes the pooled utterance-level feature vector and predicts the language label of the utterance. We use the Cross-Entropy loss function for the LID head, and this loss gets added to the main loss function with a scaling parameter.

\subsection{Multi-task Learning}
Our multi-task learning framework consists of many loss functions described below.
We use the conformer encoder output logits to the CTC layer to compute the CTC loss \boldsymbol{$\mathcal{L}_{ctc}$}. The phoneme recognition decoder PHN-DEC takes conformer encoder output and phoneme embeddings as input and generates a sequence of phoneme labels. We denote the loss calculated for the phoneme recognition task as \boldsymbol{$\mathcal{L}_{pr}$}
Similarly, the grapheme recognition decoder (GRP-DEC) is responsible for predicting the grapheme sequence for the given input audio. It takes conformer encoder output and character embeddings as input and generates a sequence of grapheme units along with the language label. We denote the loss computed by GRP-DEC as \boldsymbol{$\mathcal{L}_{gr}$}. For the language identification task, we use the utterance-level feature vector from the encoder block and compute Cross entropy loss between the predicted label and original language label. The LID loss is denoted as \boldsymbol{$\mathcal{L}_{lid}$}.
During training, we compute the weighted sum of these 4 losses \boldsymbol{$\mathcal{L}_{ctc}$}, \boldsymbol{$\mathcal{L}_{pr}$}, \boldsymbol{$\mathcal{L}_{gr}$}, and \boldsymbol{$\mathcal{L}_{lid}$} to jointly optimize the the network during training. The multi-task loss function is given by:

\begin{equation}
  \begin{aligned}
    \boldsymbol{\mathcal{L}} = \alpha  \boldsymbol{\mathcal{L}_{ctc}} + \beta \boldsymbol{\mathcal{L}_{pr}} + \gamma \boldsymbol{\mathcal{L}_{gr}} + \pi \boldsymbol{\mathcal{L}_{lid}}
 \end{aligned}
\end{equation}

The hyperparameters $\alpha$, $\beta$, $\gamma$, and $\pi$ are the scaling factors for CTC loss, phoneme recognition loss, grapheme recognition loss, and LID loss, respectively.

\subsection{Decoding}
Multilingual speech recognition involves training an acoustic model which learns to transcribe audios from multiple languages. Sometimes two or more languages will have similar acoustic units, which is very common in many Indian languages. During training, the model should be provided with the language information either to the encoder, decoder, or both. For the decoder, we investigate the condition decoding scheme in this work. The conditional decoding objective forces the GRP-DEC to predict the language label before generating the character sequence. Since the GRP-DEC decoder is autoregressive, every predicted character will be conditioned on the predicted language label. This technique helps the model to learn language identification capability. But, during the inference stage, we only use the CTC output with a 3-gram language model to generate the final hypothesis. For a given input utterance X, the final hypothesis is computed using CTC probabilities and an n-gram language model using shallow fusion as follows.

\begin{equation}
  \begin{aligned}
    Y^* =  \underset{Y \in \mathcal{Y}} {\mathrm{argmax}} \{\text{log}P_{ctc}(Y|X) + \lambda \text{ log}P_{lm}(Y)\}
 \end{aligned}
\end{equation}

Where $\lambda$ is a language model weight during beam search decoding, typically $\lambda$ is set to 1.4 is set for all the experiments. The language model is trained by combining all the text data from all six languages to create a multilingual language model. We use a 3-gram multilingual language model for all the experiments.

\begin{table*}[ht]
  \centering
  \caption{WER comparison between baseline approaches and proposed methods for test set}
  \label{tab:tasks}
    \begin{tabular}{l|c|c|c|c|c|c|c}
     \hline
      \textbf{System} & \textbf{Gujarati} & \textbf{Tamil} & \textbf{Telugu} & \textbf{Marathi} & \textbf{Hindi} & \textbf{Odiya}  & \textbf{Weighted Average}\\
      \hline
      GMM-HMM (Baseline-1)   & 28.33  & 48.81 & 47.21 & 33.22  & 69.03 & 55.78 & 46.88 \\
      TDNN-HMM (Baseline-2) & 19.27  & 33.35 & 30.62 & 22.44  & 40.41 & 39.06 & 30.73\\
      \textit{Conformer-GRP} & 21.51  & 35.78 & 34.19 & 18.01  & 27.25 & \textbf{35.56} & 28.10 \\
      \textit{Conformer-GRP+PHN} (ours) & 19.09  & 33.50 &  31.64 & 18.28  & 27.45 & 39.53 & 27.67 \\
      \textit{Transformer-GRP+PHN+LID} (ours) & 21.24  & 36.43 & 33.63 & 18.38  & 28.65 & 36.15 & 28.41\\
      \textit{Conformer-GRP+PHN+LID} (ours)  & \textbf{18.35}  & \textbf{32.18} &  \textbf{30.16} & \textbf{17.94}  & \textbf{26.20} & 38.57 & \textbf{26.70} \\
      \hline
      Relative WER reduction - BS1 & 35.22  & 34.07 & 36.11 & 47.99  & 62.04 & 30.85 & 41.04\\
      Relative WER reduction - BS2 & 4.77  & 3.5 & 1.4 & 19.87  & 35.16 & 1.25 & 10.99\\
      \hline
    \end{tabular}
\end{table*}

\begin{table*}[ht]
  \centering
  \caption{WER comparison between baseline approaches and proposed methods for blind test set}
  \label{tab:tasks}
    \begin{tabular}{l|c|c|c|c|c|c|c}
     \hline
      \textbf{System} & \textbf{Gujarati} & \textbf{Tamil} & \textbf{Telugu} & \textbf{Marathi} & \textbf{Hindi} & \textbf{Odiya}  & \textbf{Weighted Average}\\
      \hline
      TDNN-HMM (baseline) & 26.15  & 34.09 & 31.44 & 29.04  & 37.20 & 38.46 & 32.73 \\
      EthereumMiner & 20.11  & 28.52 & 26.08 & 20.15  & 17.54 & 19.99 & 22.06  \\
      Uniphore & 23.48  & 18.58 & 29.26 & 18.58  & 24.35 & 28.67 & 23.82 \\
      Ekstep & 30.65  & 27.2 & 22.43 & 12.24  & 39.74 & 27.1 & 26.56 \\
      \textit{Conformer-GRP+PHN+LID} (ours) & 34.79  & 37.95 & 34.79 & 96.75  & 20.74 & 36.07 & 43.40 \\
      \hline
    \end{tabular}
\end{table*}

\section{Dataset and Experiments}
We conduct our experiments on the data set released by Microsoft and Navana Tech as part of a special session on “Multilingual and code-switching ASR challengees for low resource Indian languages in INTERSPEECH 2021”. The challenge has two main subtasks, 1) multilingual speech recognition and 2) code-switching speech recognition. In this work, focus on subtask-1. The subtask-1 dataset includes training and testing data for six Indian languages - Telugu, Tamil, Gujarati, Marathi, Hindi, and Odiya. The dataset comes from many domains, including Stories, healthcare, agriculture, finance, and general topics. Some languages, such as Odiya, Marathi, and Hindi, are sampled at 8KHz, whereas Telugu, Tamil, and Gujarati are sampled at 16Khz. The dataset also comes with a lexicon for each language. The dataset contains train, and test for each language. Blind test evaluations are conducted as part of the challenge. Telugu, Tamil, and Gujarati, Each contains 40hrs of training data and 5hrs test data, whereas Hindi, Marathi, and Odiya contain ~95Hrs of training ~5hrs of test data. We report all our results on both test sets and blind test data.

The conformer encoder contains 12 conformer blocks with two initial convolution layers for sub-sampling. Each conformer block contains eight attention heads with 512 units for the attention head and 2048 hidden units for the pointwise feed-forward layer. It also uses the positional encoding layer to learn positional information of the speech frames. The vocabulary of PHN-DEC consists of 68 phonemes excluding $<$unk$>$, $<$blank$>$,$<$space$>$ and $<$sos/eos$>$ tags. The vocabulary of GRP-DEC consists of 341 graphemes from all the languages combined without extra tags. We augment the vocabulary $<$unk$>$, $<$blank$>$,$<$space$>$ , $<$sos/eos$>$ and six language labels. The language labels are [TE] (Telugu), [TA](Tamil), [GU](Gujarati), Marathi [MA], Hindi [HI], and Odiya [OD]. Our final grapheme vocabulary contains 351 tags in total.
Our model hyperparameters $\alpha$, $\beta$, $\gamma$, and $\pi$ are set to 0.3, 0.5, 0.5, and 10.0 respectively. We use 40-dimensional Mel-filterbank features extracted with a 25ms frame window, with 10ms frameshift. The training dataset is augmented with speech perturbation and SpecAugment [27]. For Language model integration, we use a 3-gram multilingual language model. We set $\lambda$ to 1.4 for all the experiments and beam size to 20. We use Adam optimizer [28] with an initial learning rate of 0.001, and the learning rate is decayed using a learning rate scheduler based on validation loss. We train our model up to 30 epochs with a batch size of 20.

\section{Results}
In this section, we briefly discuss results of our experiments. We first compare our proposed approach with the baseline methods and the effect of the number of layers in the phoneme recognition decoder. We finally show our results on the blind test set and compare it with the top 3 winning systems.

\subsection{Comparison with baseline methods}
For the baseline, we use two systems provided by the challenge organization committee, GMM-HMM (baseline-1) and TDNN-HMM (baseline-2). Both baseline-1 and baseline-2 are hybrid speech recognition models, and the WER for the test data for these models is given in Table 1. The system \textit{Conformer-GRP+PHN+LID} represents our proposed model. It contains a 12 layer conformer encoder, six transformer layers at the decoder for grapheme recognition, and three transformer layers for phoneme recognition. It also contains a language classifier with two linear layers of size 128 and 64. It can be seen that our proposed approach outperforms the baselines by a large margin. We obtain 41.01\% average reduction in WER (relative) compared to baseline-1 and 10.99\% reduction compared to baseline-2. Injecting language information to the multilingual ASR model is an important factor, and it can be provided at both the encoder and decoder parts of the model. The experiment \textit{Conformer+GRP+PHN+LID} provides the language information at the encoder and decoder both, whereas \textit{Conformer+GRP+PHN} provides language information only at the decoder. It can be seen that \textit{Conformer+GRP+PHN+LID} obtains better performance compared to the system which has no language classifier (\textit{Conformer+GRP+PHN}).
The blind test set evaluations are shown in Table 2. We can observe that our model poorly on Marathi due to mismatched channel conditions.

\subsection{Transformer vs Conformer}
To compare the transformer and conformer models, we set up two experiments known as \textit{Transformer-GRP+PHN+LID}, which uses 12 layers of Transformer blocks at the encoder and six transformer layers at the decoder GRP-DEC, and three transformer layers at the decoder PHN-DEC. It can be seen that from Table 1, the conformer-based approach obtains a 6\% relative improvement in WER compared to the transformer-based approach due to its nature of capturing both local and global variations in the audio.

\subsection{Effect of number of layers in PHN-DEC}
In order to find the effect of the number of decoder layers on the performance, we train five different models with 1,2,3,4 and 6 transformer layers at PHN-DEC decoder while keeping six transformer layers for GRP-DEC. We calculate both the Character Error Rate (CER) and Word Error Rate (WER) of each model on the test data. The WER/CER with the increasing number of decoder layers at PHN-DEC, as shown in Figure 2. We observe that keeping three decoder layers for PHN-DEC and six for GRP-DEC obtains the best WER/CER.

\begin{figure}[t]
  \centering
  \includegraphics[width=\linewidth]{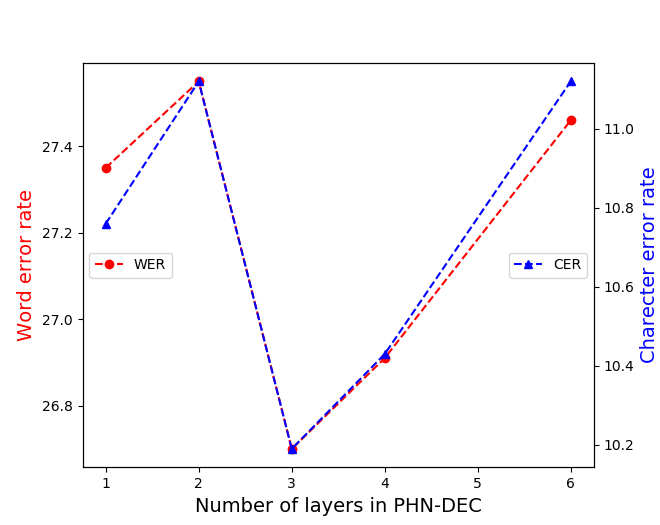}
  \caption{WER/CER w.r.t number of decoder layers in PHN-DEC}
  \label{fig:speech_production}
\end{figure}
 
\section{Conclusions}
Multilingual speech recognition is one of the challenging problems in speech processing. In this work, we propose a dual-decoder approach for multilingual speech recognition tasks for three Indian languages Gujarati, Tamil, and Telugu. We propose to use a conformer encoder with two parallel transformer decoders for the multilingual speech recognition task. We use phoneme recognition and language identification as auxiliary tasks to learn shared feature representation in the multi-task learning framework. We show that our dual-decoder approach obtains more than a 41.01\% and 10.99\% average relative reduction in WER compared to baseline-1 and baseline-2, respectively. We also show that our proposed conformer encoder systems obtain a significant reduction over the transformer-based systems. 
\bibliographystyle{IEEEtran}

\end{document}